\documentclass[preprint,12pt]{elsarticle}

\usepackage{amssymb}
\usepackage{amsmath}

\usepackage{booktabs}
\usepackage{algorithm}
\usepackage{algpseudocode}
\usepackage[table]{xcolor}
\usepackage[super]{nth}
\usepackage{subcaption}

\usepackage{graphicx}
\usepackage{tikz}

\newcommand*\circled[1]{\tikz[baseline=(char.base)]{
            \node[shape=circle,draw,inner sep=0.5pt] (char) {#1};}}

\makeatletter
\def\ps@pprintTitle{%
 \let\@oddhead\relax
 \let\@evenhead\relax
 \let\@oddfoot\relax
 \let\@evenfoot\relax}
\makeatother

\begin{document}

\begin{frontmatter}



\title{Towards a Trustworthy Anomaly Detection for Critical Applications through Approximated Partial AUC Loss}



\author[label1]{Arnaud Bougaham} 
\author[label1]{Beno\^it Fr\'enay}
\affiliation[label1]{organization={University of Namur - NaDI - HuMaLearn Lab - Faculty of Computer Science},
            addressline={Rue Grandgagnage 21}, 
            city={Namur},
            postcode={B-5000}, 
            country={Belgium}}

\begin{abstract}
Anomaly Detection is a crucial step for critical applications such in the industrial, medical or cybersecurity domains. These sectors share the same requirement of handling differently the different types of classification errors. Indeed, even if false positives are acceptable, false negatives are not, because it would reflect a missed detection of a quality issue, a disease or a cyber threat. To fulfill this requirement, we propose a method that dynamically applies a trustworthy approximated partial AUC ROC loss (\textit{tapAUC}). A binary classifier is trained to optimize the specific range of the AUC ROC curve that prevents the True Positive Rate (TPR) to reach 100$\%$ while minimizing the False Positive Rate (FPR). The optimal threshold that does not trigger any false negative is then kept and used at the test step. The results show a TPR of 92.52\% at a 20.43\% FPR for an average across 6 datasets, representing a TPR improvement of 4.3$\%$ for a FPR cost of 12.2\% against other state-of-the-art methods. The code is available at https://github.com/ArnaudBougaham/tapAUC.

\end{abstract}




\end{frontmatter}




\section{Introduction}
\label{Introduction}


In recent years, modern organizations have taken advantage of artificial intelligence (AI) advances to optimize their processes. In the same time, explainable AI (XAI) techniques have been developed to improve the interpretability and explanability in these AI systems. However, this is sometimes not sufficient in critical domains, where strong trustworthiness constraints are required. A group of AI experts for the European Commission released the Ethics Guidelines for Trustworthy AI where some of the proposals are to improve, among other things, the robustness, the transparency or the accountability \cite{Bouadi2024trust}. This paper seeks to focus on these characteristics of trustworthiness for critical applications.

The current \nth{4} industrial revolution brings many new techniques thanks to AI, from predictive maintenance \cite{nunes2023challenges,zonta2020predictive} to production processes optimization \cite{weichert2019review}, including sales forecasting \cite{cheriyan2018intelligent}. The medical domain is also quickly adopting these new techniques, from hospital patients journey optimization \cite{el2021machine} to doctor's visit reporting \cite{falcetta2023automatic}, or even health lifestyle recommendations in relation to a suspected pathology \cite{chatterjee2023ai}. AI techniques can also contribute in these fields to diagnose problems, through anomaly detection (AD) systems \cite{wen2024survey,li2024intelligent,abou2024generative}. Indeed, the industrial and medical sectors share many common features, when it comes to detect a quality issue or a disease. Among them, we identify the decisions based on image processing techniques, the lack of abnormal data yielding to imbalanced datasets, or the importance to automate painful and time-consuming tasks. But one of the most important concern is the necessity to avoid any critical missed detection, while keeping a small rate of false alarms \cite{bougaham2025industrial,yang2021all}. Indeed, in critical industries like automotive, health, nuclear or aerospatial, a product test coverage that fails to meet the quality specifications would bring dramatic consequences. A similar reasoning is applicable for the medical domain, where a specialist would not detect symptoms heralding a disease that has to be treated quickly. Therefore, a reasonable scheme to deal with such constraints is to strengthen the decisions that a human expert has to take, with an AI-assistant tool such as a normal/abnormal classifier.
This statement motivates this study, to assist the operators or specialists in their decision making. 

In this context, one of the most challenging step is to build a binary classifier for anomaly detection that is accurate, trustworthy and transparent. Each of these characteristics would encourage the business experts to adopt the technology and would increase its decision-making quality. This implies that such a model has to be trained with the objective to yield an asymmetric confusion matrix, with zero false negatives (ZFN) and a limited false positive (LFP) ratio. In other words, from the training to the inference step, the method has to be tailored such that it gives a larger importance to positive (abnormal) instances than negative (normal) ones. In addition, the tool would have to give a score-based degree of confidence, in order to guide the expert, as well as to bear full responsibility for his final decision.

In the critical or imbalanced learning literature, many methods exist to deal with the constraint, namely oversampling the minority class \cite{ahsan2022comparative} or undersampling the majority class \cite{fernandez2018learning}, giving a larger weight to the loss when dealing with the minority class \cite{cui2019class}, or adapting the threshold after the training phase so that all positives are classified as true ones \cite{bougaham2021ganodip}. All these benefits come with drawbacks. Among them, there are the sacrifice of informative data or the difficulty to generate synthetic ones. We can also cite the need to assign the right cost or the absence of the constraint during the training step. This study considers all of these drawbacks and tackles them by designing a loss function in a specific method called \textit{tapAUC} (trustworthy approximated partial Area Under the Curve) that incorporates the quality constraint during the optimization, without giving any specific weight for the unacceptable classification errors. This work is the continuation of our preliminary one \cite{bougaham2023composite}, where an industrial partner reinforces its quality strategy for an automotive PCBA (Printed Circuit Board Assembly) production line. The objective is to prevent any defect to be missed while maintaining an LFP ratio, and show how the method can generalize to other domains sharing the same concern. To do so, an approximation of the Area Under the Curve of the Receiver Operating Characteristic (AUC ROC) metric is considered as the loss function of a feed-forward neural network classifier. The positive instances, and a partial selection of the negative ones, help focusing on the region of interest to optimize, namely the one that prevents a full TPR at the minimum FPR possible. Our main contributions are the following:
\begin{itemize}
\item Use an approximated pAUC loss for a trustworthy anomaly detection.
\item Dynamically focus on negative instances responsible for critical errors.
\item Run experiments on industrial, medical and cybersecurity datasets.
\item Compare the results with other state-of-the-art methods.
\item Build an uncertainty interval to better engage the expert responsibility.
\item Discuss the method in the context of real-world applications.
\end{itemize}
This study is organized as follows. Section~\ref{Related Works} introduces the necessary related works to fully understand the previous approaches and the subsequent concepts. Then Section~\ref{Method} details the method proposed, from the customized loss presentation to the algorithm description. After that, Section~\ref{Evaluation} presents the experiments and the results are discussed in Section~\ref{Discussion}. Finally, Section~\ref{Conclusion} summarizes the study and proposes future works.


\section{Related Works}
\label{Related Works}


Taming a binary classifier is an old game that researchers have been playing for many years. The critical applications, where it is unacceptable that instances belonging to a risky group would be misclassified, push them to raise methods where the training step penalizes more false negatives that false positives. This is also the case in the imbalanced literature where the critical class is often the minority one. In this context, cost-sensitive mechanisms were developed, where the minority class loss is adapted to compensate its under-representation \cite{cui2019class,ross2017focal,cao2019learning}. In a critical application setting, it helps the classifier to better classify the positive class and avoid to generate false negatives. The drawback here is that there is no guarantee that the methods assign the right cost, yielding a biased loss function. In \cite{chawla2002smote} and \cite{ahsan2022comparative}, the authors proposed to oversample the minority class, giving more synthetic examples to the classifier that learns better from this underpopulated class. Here, the challenge is to make these generated examples as close as possible to the real distribution, and it is not easy to fulfill this requirement (specifically for high dimensional data). For similar reasons, the authors of \cite{drummond2003c4} and \cite{fernandez2018learning} decided to undersample the majority class, requiring to sacrifice informative data that would eventually be interesting to keep for the classification performance. Also, some studies \cite{bougaham2021ganodip,roth2022towards} selected the threshold that does not generate any false negatives, after training. The drawback here is that the model does not consider the zero false negative constraint during the optimization, and its performance is not optimal in that sense. 

Another way to focus on the TPR is to approximate the AUC ROC curve and directly use it as the loss function, instead of the traditional binary cross entropy. Indeed this AUC ROC metric shows the True Positive Rate (TPR) against the False Positive Rate (FPR) for different thresholds, making it possible to optimize both together. In the beginning of the \nth{21} century, the authors of \cite{yan2003optimizing} laid the foundations by introducing an approximation of the AUC ROC metric, through the Wilcoxon-Mann-Whitney statistic \cite{mann1947test}. They showed that it is relevant to train a classifier the same way it will be evaluated, with the AUC metric. They proposed a differentiable surrogate loss function to make it possible. To focus on a specific part of the AUC, a one-way partial AUC loss has been developed \cite{dodd2003partial}, being a surrogate loss that only considers a portion of the AUC curve where the FPR is bounded between a low and a high value. A one-way and two-way partial AUC loss methods are then proposed in \cite{zhu2022auc}, taking into account a portion of the FPR as well as a portion of the TPR. The goal is to simulate a TPR and a FPR threshold to optimize a very specific part (the top left one) of the AUC ROC. They used a weakly convex optimization algorithm to train the classifier. However, the method focuses on a larger portion of the AUC ROC curve than the specific one that prevent any false negatives at the full TPR regime. After this literature survey, one can observe that an algorithm tailored to optimize the FPR at the very specific 100\% TPR setting, during training, is still lacking. The next section \ref{Method} precisely describes such a new method.

\section{The tapAUC Method}
\label{Method}

After having introduced the context and described the related works, the proposed method \textit{tapAUC} (trustworthy approximated partial AUC) is detailed in this section.

\subsection{Zero False Negatives under Limited False Positive Rate}
\label{ZFN_LFP}

\begin{figure}[t!]
\centering
\begin{tikzpicture}
    \node[anchor=south west, inner sep=0] (image) at (0,0) 
          {\includegraphics[width=1\textwidth]{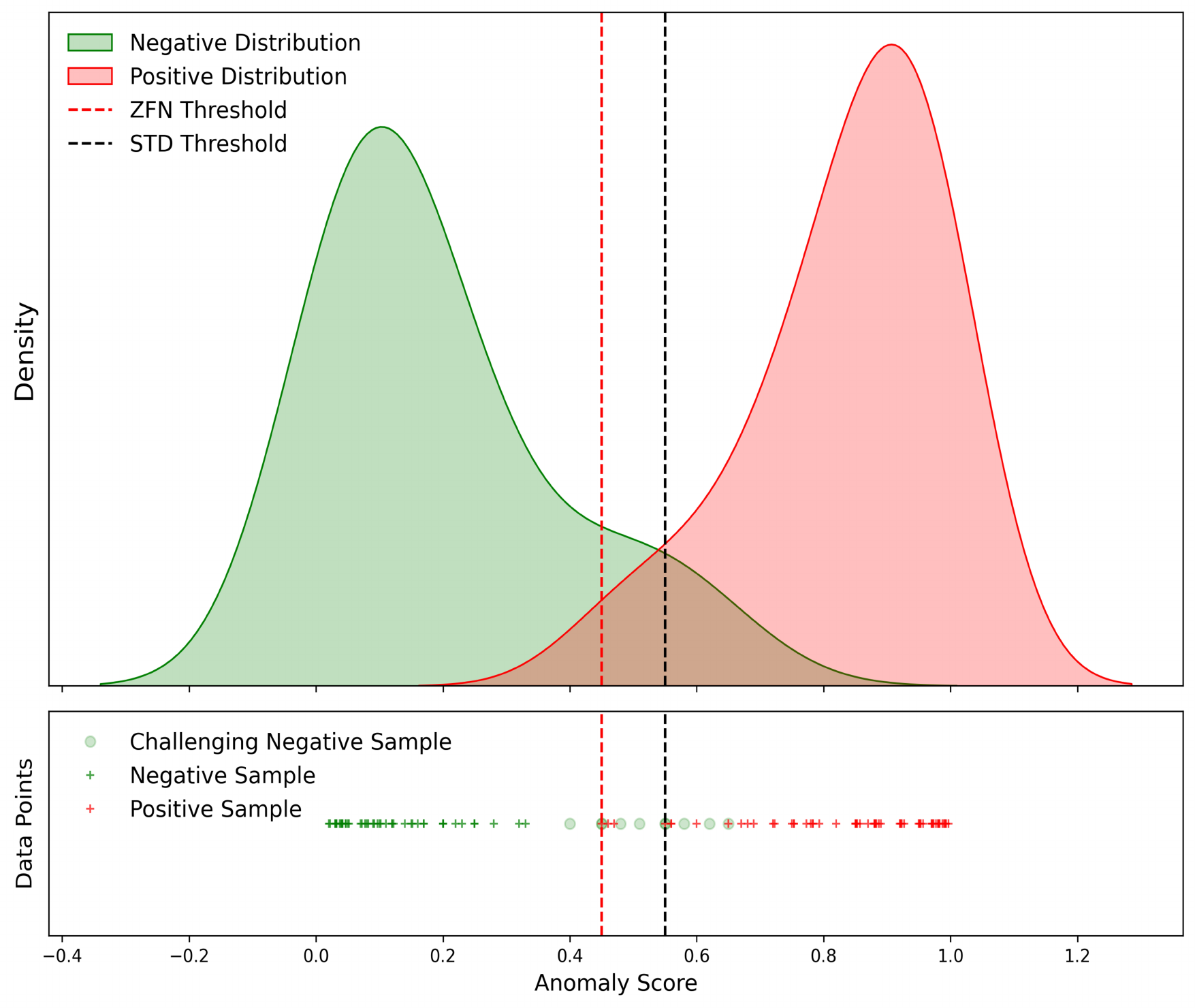}}; 
    \begin{scope}[x={(image.south east)},y={(image.north west)}]
        \node[draw, circle] at (0.35,0.125) {1};
        \node[draw, circle] at (0.53,0.125) {2};
    \end{scope}
\end{tikzpicture}
\caption{Anomaly score data points and distribution example of negative (in green) and positive (in red) instances. Unlike the threshold that yields the maximum accuracy (STD Threshold dashed black vertical line), the one that reaches 100\% TPR (ZFN Threshold dashed red vertical line) is highly influenced by the positive and the most challenging negative data. Lower score for these challenging negatives and higher score for positives would end up with a full TPR classifier, with limited FPR.}
\label{discrimination}
\end{figure}

\begin{figure}[p]
\centering
\includegraphics[width=.75\linewidth]{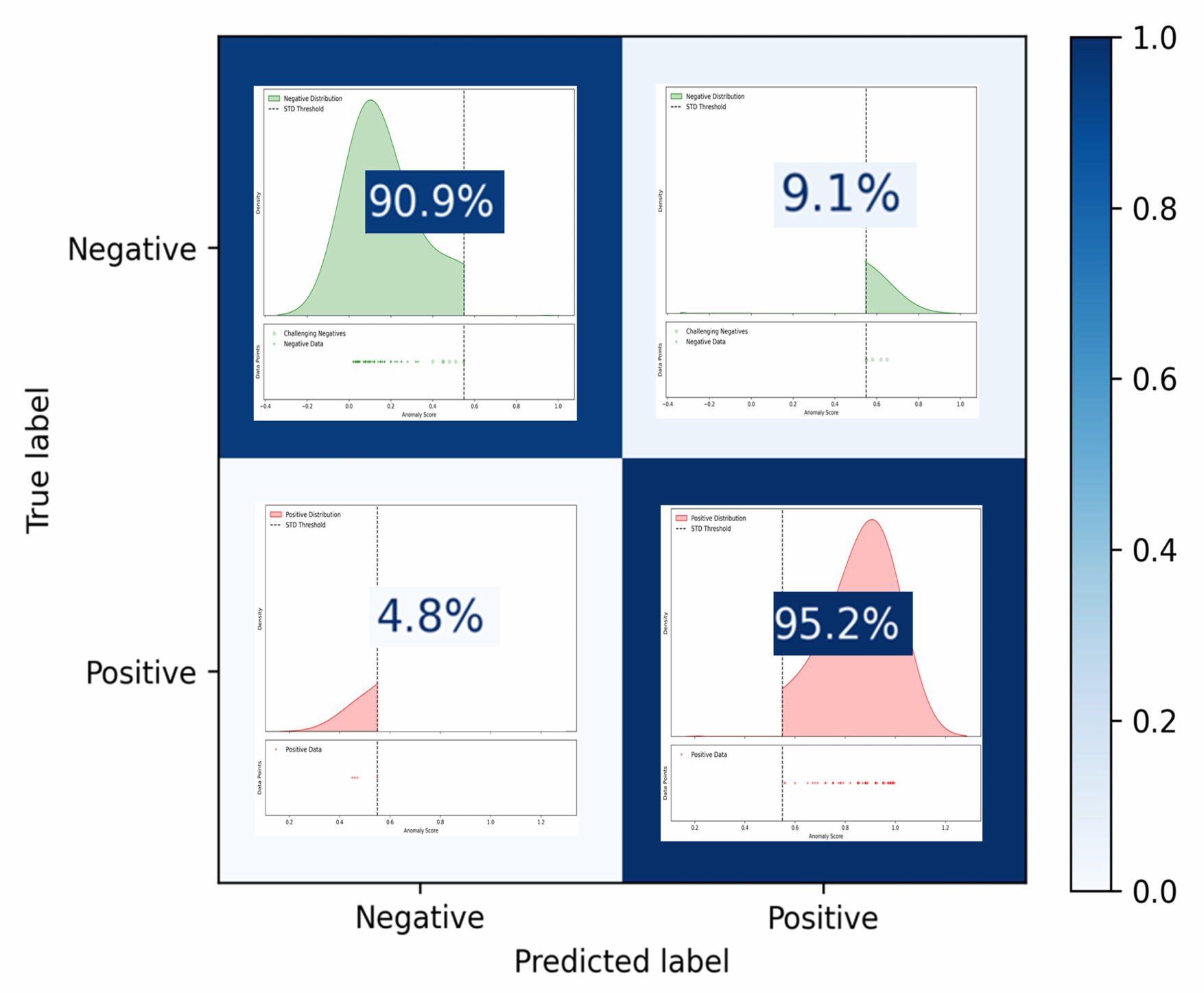}
\caption{Confusion Matrix that maximizes the accuracy.}\label{Confusion_Matrix_STD}
\vspace{20pt}
\centering
\includegraphics[width=.75\linewidth]{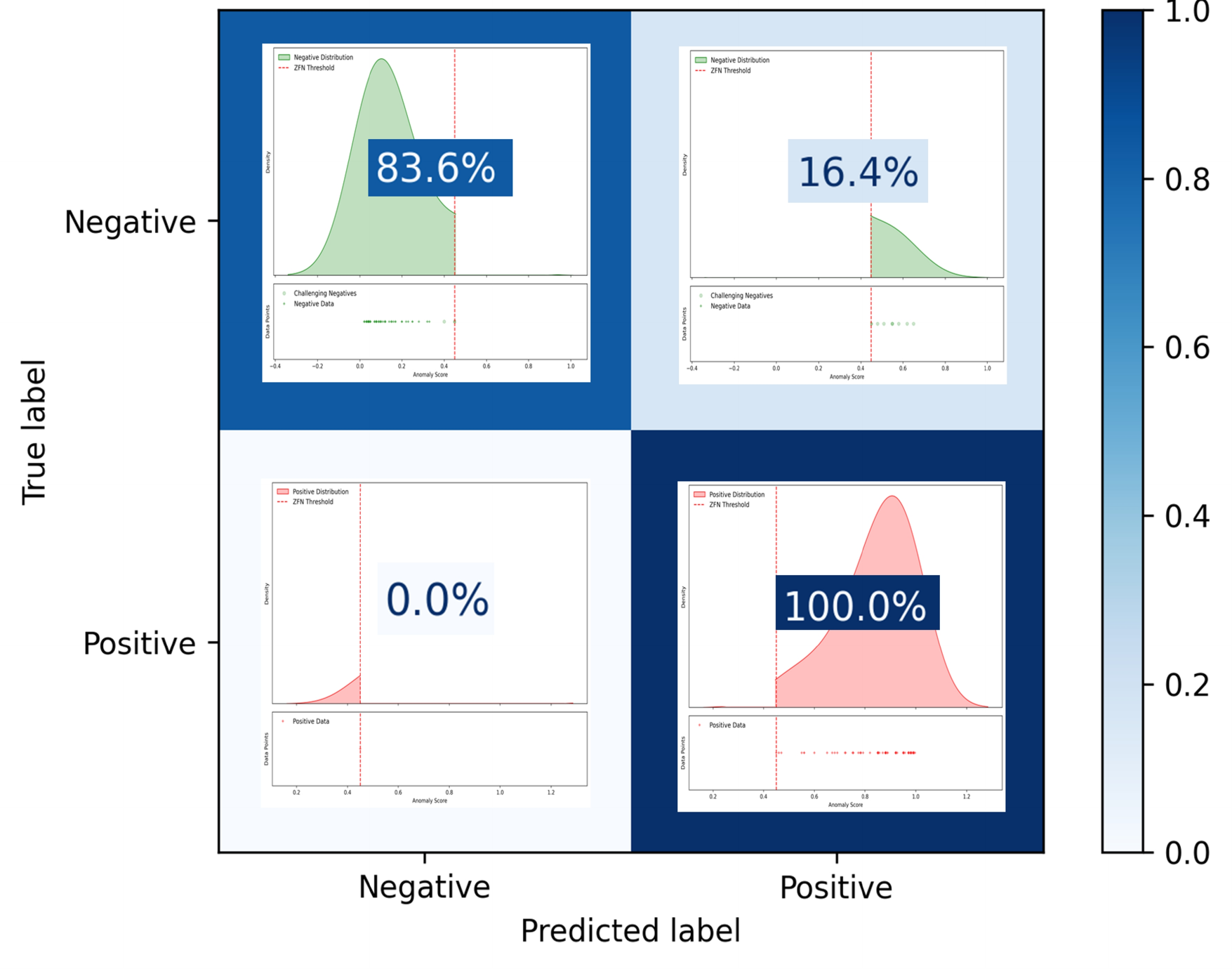}
\caption{Confusion Matrix that guarantees ZFN under LFP.}\label{Confusion_Matrix_ZFN}
\end{figure}

The main objective is to train a binary classifier able to tackle the critical application problem. To do so, we will consider the True Positive Rate (TPR) and the False Positive Rate (FPR):
\begin{equation}
    \text{TPR} = \frac{TP}{TP + FN},\nonumber
\end{equation}
\begin{equation}
    \text{FPR} = \frac{FP}{FP + TN}\nonumber
\end{equation}
where $TP$ is the set of True Positives, $FN$ the False Negatives, $FP$ the False Positives and $TN$ the set of True Negatives.
Note that 100\% TPR is equivalent to 0\% False Negative Rate (FNR) because $TPR=1-FNR$ with:
\begin{equation}
    \text{FNR} = \frac{FN}{FN + TP}\nonumber
\end{equation}
Therefore, a classifier that reaches a full TPR does not generate any false negatives. This zero false negative (ZFN) setting makes it possible to detect all the positives, which is a key feature for a critical application such in the industrial, the medical or the cybersecurity domain. Figure~\ref{discrimination} shows this statement through an example.
During training, the model predicts the scores to associate the class that each instance should belong to, where the normal class is 0 and the abnormal class is 1. These scores are the data points shown in the bottom part of the figure. We end up with anomaly scores distributed between 0 and 1 for both classes, shown in the top part of the figure, that have to be separated by a threshold to classify each data as negative or not. The ones with an anomaly score below the threshold will be classified as negative (normal), and the ones above will be classified as positive (abnormal). After that, the number of true and false positives can be calculated, depending on the threshold chosen. This threshold value is very important when dealing with critical applications. Indeed, the one that classify all the positives correctly has to be selected, in order to generate zero false negative (ZFN) and guarantee that no detection would be missed. The other constraint is to reduce the number of false alarms, in order to use this tool without permanently alarming the user about anomalies that do not exist. This limited false positives (LFP) ratio requires the classifier to minimize the FPR. In Figure~\ref{discrimination}, the threshold that corresponds to these constraints is the red dashed line. It is set at the exact left border of the positive distribution, to prevent any additional false positives, while detecting all the positives. In other words, its optimal value for our critical applications is the minimum anomaly score of the positive set. In order to observe how the true and false positives are distributed through the threshold configurations, Figure~\ref{Confusion_Matrix_STD} shows the confusion matrix for the standard STD threshold that maximizes the accuracy, and Figure~\ref{Confusion_Matrix_ZFN} shows the one that is of interest in our context, being the ZFN threshold.
For a better understanding, the part of the distribution concerned is displayed for each cases. On can observe that the confusion matrix with the ZFN threshold in Figure~\ref{Confusion_Matrix_ZFN} yields 0 false negative (no true positive predicted as negative), which is the not the case with the STD threshold in Figure~\ref{Confusion_Matrix_STD}. The drawback is that the ZFN FPR (true negative predicted as positive) is higher than the STD one. Despite this statement, this ZFN setting is exactly the one that will be exploited during the classifier training. 

\subsection{Partial AUC ROC Curve}
\label{pAUC}

The objective of the proposed \textit{tapAUC} method is to exploit this ZFN under LFP mechanism, during training. An approximation of the AUC ROC curve, which displays the TPR against the FPR for all the possible thresholds, is therefore chosen as the loss function. Commonly used to assess the classifier performance after training, this metric is customized and used in a dedicated algorithm, to bring the class importance knowledge during the model optimization. Using this loss during the training procedure improves the performance because it directly targets what we aim for, instead of a traditional binary cross-entropy as it is usually used. But it also offers a direct access to the TPR and the FPR metrics, where we can slightly arrange the loss to keep improving the FPR only in the 100\% TPR setting. Indeed, instead of considering all the data in this approximated AUC loss, only a part of the negative instances and all the positive ones will be selected. The loss function thus becomes a partial AUC (pAUC) loss. The targeted negatives subset is the one that generates high anomaly scores because its instances are difficult to correctly classify. And these instances are the most interesting for the model to learn about the task, much more than easy ones that do not help reducing the overlap region during training.

This pAUC loss brings the necessary information to the classifier so that it modifies its parameters to better classify a specific cut-off region of interest. Indeed, instead of improving all the unnecessary regions of the AUC ROC, with the negatives that do not influence the ZFN threshold placement (corresponding to the region \circled{1} in Figure~\ref{discrimination}), this approach reduces the focus at the more challenging instances that can result a higher anomaly score than some positives (region \circled{2} in Figure~\ref{discrimination}). Iteration after iteration, the algorithm computes the loss for all the positive instances and only this subset of the negative ones. 
The lower the anomaly score of the challenging negative instances, the better the discrimination while considering the specific 100\% TPR threshold.

\begin{figure}[p]
\centering
\includegraphics[width=1\linewidth]{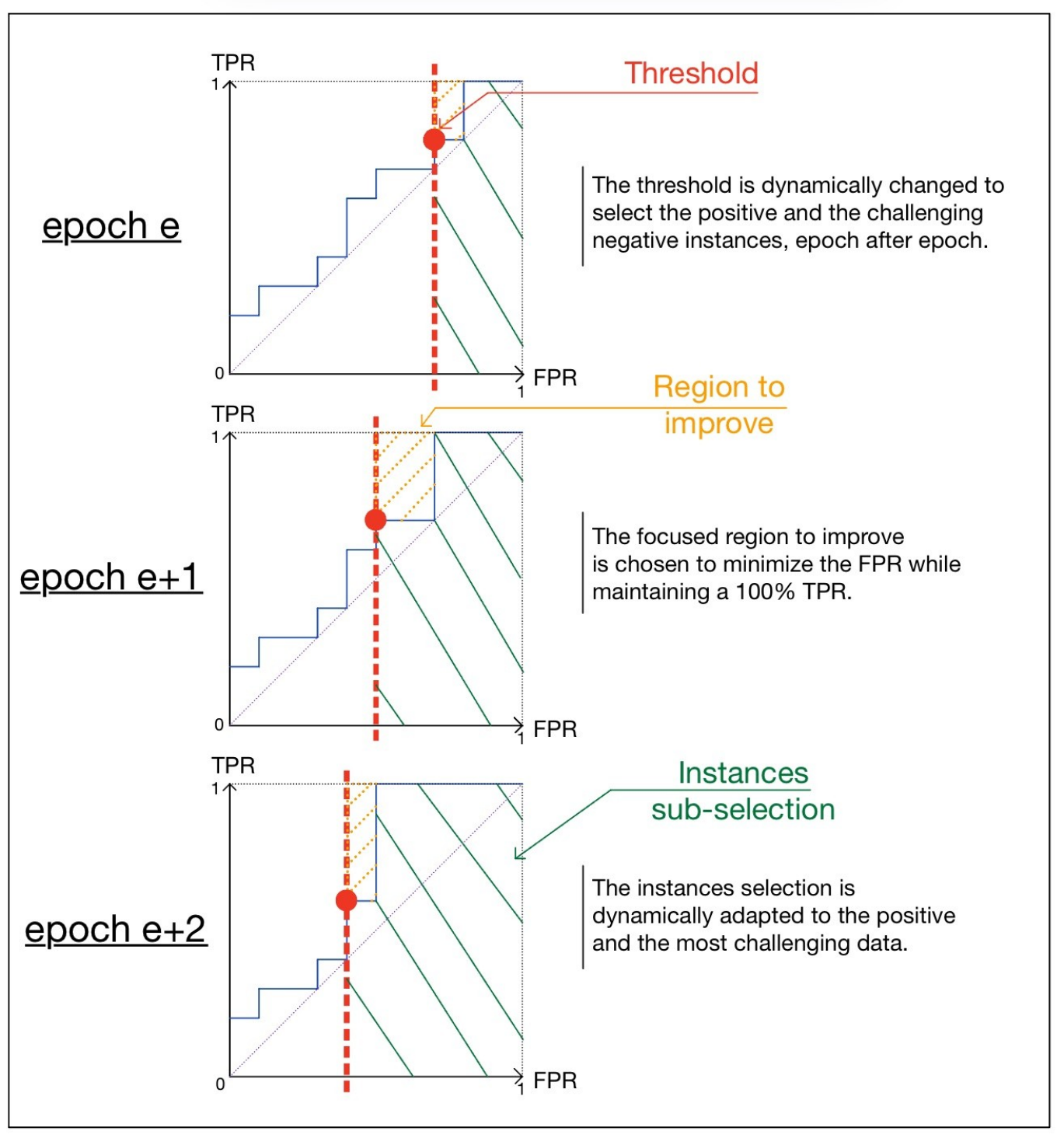}
\caption{During the classifier training, epoch after epoch, the negative instances that just prevent a 100\% TPR and all the positive ones are selected. The customized loss function approximates the partial AUC ROC curve, where the focused improved region helps minimizing the FPR while maintaining the TPR at 100\%.}\label{AUC}
\end{figure}

Figure~\ref{AUC} shows this mechanism in the partial AUC view. The objective is to lower the score of the challenging negative data close to the positive one. If so, negative and positive distributions could be better separated, by lowering the threshold. As explained in the top part of Figure~\ref{AUC}, the threshold is selected so that all the positive and almost all negative instances are well classified. Only the few misclassified negatives are considered, being the most challenging ones. The middle part of the figure shows that, while the model gets optimized, this threshold will change (at the next epoch) to always follow this strategy and target a specific region to improve. For our application, it has therefore to be placed at $TPR\approx1$, to let the opportunity for the model to keep optimizing the very small part that do not yet reach 100\%, while not focusing on the other uninteresting regions. The bottom part of the figure explains how the subsample selection evolves epoch after epoch. We end up with an optimal model in the context of a minimal FPR and full TPR, particularly sensitive with challenging negative data, by getting rid of the other easily classified ones.

\subsection{Loss Function}
\label{Loss Function}

The standard way to discriminate classes is to train a model based on the binary cross entropy (BCE) loss. Indeed, this loss is well suited to handle two classes and to backpropagate its gradient while training. Nevertheless, it does not take into account the nature of the classification error, whatever it is a false positive or a false negative. This is why an approximation of the pAUC loss is needed. The objective is to perform pairwise comparisons of the anomaly scores through a surrogate squared hinge loss, for the positive instances and the more challenging negative ones. Consider a dataset \( \mathcal{D} = \{X, Y\} \), with \( N \) instances of input features \( X = \{x_1, \ldots, x_N\} \), and its corresponding binary labels \( Y = \{y_1, \ldots, y_N\} \) (\( y_p = 1 \) for positive instances and \( y_n = 0 \) for negative ones). A feed-forward neural network classifier \( f_\theta: \mathbb{R}^D \to \mathbb{R} \) maps the feature space to anomaly scores \( s = f_\theta(x) \), where \( \theta \) denotes the model parameters. The objective is to train \( f_\theta \) by minimizing a loss function \( \mathcal{L} \), which ensures that positive (abnormal) scores \( s_p \) are separated and higher from negative (normal) scores \( s_n \), while focusing on challenging instances. Two subsets are defined based on the output scores:
\[
\mathcal{P} = \{s_p = f_\theta(x) \,|\, y_p = 1\}, \quad \mathcal{N} = \{s_n = f_\theta(x) \,|\, y_n = 0\}.
\]
The AUC loss we want to minimize is defined as:
\begin{equation}
\mathcal{L}_\text{AUC} = \frac{1}{|\mathcal{P}| |\mathcal{N}|} \sum_{p \in \mathcal{P}} \sum_{n \in \mathcal{N}} \mathbf{1}(s_p < s_n),
\label{Lauc}
\end{equation}
where \( \mathbf{1}(\cdot) \) is the indicator function. To enable gradient-based optimization, the indicator function is replaced by a differentiable approximation, namely a squared hinge loss function:
\begin{equation}
\mathcal{L}_\text{aAUC} = \frac{1}{|\mathcal{P}| |\mathcal{N}|} \sum_{p \in \mathcal{P}} \sum_{n \in \mathcal{N}}  \max \left(0, \left(s_n + \gamma - s_p\right)\right)^2
\label{LaAUC}
\end{equation}
where \( \gamma > 0 \) is a margin separating positive and negative scores. As explained in Section~\ref{pAUC}, only a portion of the ROC curve corresponding to the challenging negative instances and all the positive ones are considered. The partial AUC (pAUC) restricts the evaluation to a specific FPR range \([ \alpha, 1 ]\) that allows this restriction. The subset of negative scores \( \mathcal{N}_{\alpha} \) is defined by:
\begin{equation}
    \mathcal{N}_{\alpha} = \mathcal{N}_{sorted}[ : \lfloor \alpha |\mathcal{N}| \rfloor],
\label{Ssorted}
\end{equation}
with $\mathcal{N}_{sorted}$ being the negative set \( \mathcal{N} \) sorted by the scores in the descending order. The trustworthy approximated pAUC loss $\mathcal{L}_\text{tapAUC}$ is expressed as:
\begin{equation}
\mathcal{L}_\text{tapAUC} = \frac{1}{|\mathcal{P}| |\mathcal{N}_{\alpha}|} \sum_{p \in \mathcal{P}} \sum_{n \in \mathcal{N}_{\alpha}}  \max \left(0, \left(s_n + \gamma - s_{p}\right)\right)^2
\label{LtapAUC}
\end{equation}
In order to warmup the classifier training, $\mathcal{N}_{\alpha}$ is built as an adaptive subset where $\alpha$ = 0 during this period (equivalent to Equation~\ref{LaAUC}), and \( 0 < \alpha < 1 \) otherwise.
It helps the model to select the most challenging examples that will be the starting point for the post-warmup step. Then, the negative data with the lowest prediction scores will be excluded from the $\mathcal{N_{\alpha}}$ subset. The more challenging negatives added to the positives influence directly the false negative rate because they condition the threshold value to place in the 100\% TPR setting. 
Iteration after iteration, the negative subset is updated, in order to dynamically track the small top left corner that is of interest (the one that just prevents 100\% TPR), and the partial AUC loss $\mathcal{L}_\text{tapAUC}$ decreases until there is no possibility anymore. At the end, the model is trained with its parameter values selected to get the best FPR while having 100\% TPR for the train set. Once the classifier is trained with this method, the 100\% TPR threshold is kept (being the smallest positive anomaly score) for evaluation. Then, the test set is inferred and metrics such as the accuracy, the TPR and the FPR are computed based on the specific threshold. The expectation is that the TPR (under a reasonable FPR) is improved compared to a different method where this focus is not done.

\subsection{Algorithm}
\label{Algorithm}

The \textit{tapAUC} method is summarized in the pseudo-code Algorithm \ref{Pseudo-code}. The training loop makes the classifier optimize its parameters based on Equation~\ref{LtapAUC}. Before the warmup period, the loss considers the full train set, with all the negative and positive instances. After this period, only a partial set of negatives, combined to all positives, is dynamically considered, epoch after epoch. Once the classifier trained, the threshold that gives 100\% TPR on the train set is kept, and used to compute the metrics on the test set.

\begin{algorithm}
\caption{Training and Testing tapAUC Model}
\label{Pseudo-code}
\begin{algorithmic}[1]
\Procedure{TrainAndTest}{$train\_loader, test\_loader$}
    \State Initialize $model$, optimizer, $e\_total$, $e\_warmup$, $\alpha$
    \For{$epoch = 0$ to $e\_total$}
        \State Set $model$ to training mode
        \For{each batch $(X\_batch, y\_batch)$ in $train\_loader$}
            \State $y\_pred \gets model(X\_batch)$
            \If{$epoch \geq e\_warmup$}
                \State Sort $y\_pred$ scores for $N$ (negatives) and $P$ (positives)
                \State Create $N_\alpha$  with the $\alpha$ highest scores of $N$
                \State Combine $N_\alpha$ and $P$ to form $y\_batch_\alpha$ with $y\_pred_\alpha$ scores
                \State $loss \gets SquaredHingeLoss(y\_pred_\alpha, y\_batch_\alpha)$
            \Else
                \State $loss \gets SquaredHingeLoss(y\_pred, y\_batch)$
            \EndIf
            \State Perform backward pass and optimize
        \EndFor
        \EndFor
        \State $threshold_{ZFN} \gets$ lowest prediction score of $P$
        
        \State Set $model$ to evaluation mode
        \For{each batch $(X\_batch, y\_batch)$ in $test\_loader$}
            \State $y\_pred \gets model(X\_batch)$
            \State Compute $Accuracy, TPR, FPR$ based on $threshold_{ZFN}$
        
    \EndFor
    \State \Return $threshold_{ZFN}, Accuracy, TPR, FPR$
\EndProcedure
\end{algorithmic}
\end{algorithm}

\section{Evaluation}
\label{Evaluation}

This section shows the evaluation performed to assess the proposed \textit{tapAUC} method. The datasets, the experimental setup and the results are described.

\subsection{Datasets}
\label{Datasets}



\begin{figure}[h!]
\centering
\includegraphics[width=1\linewidth]{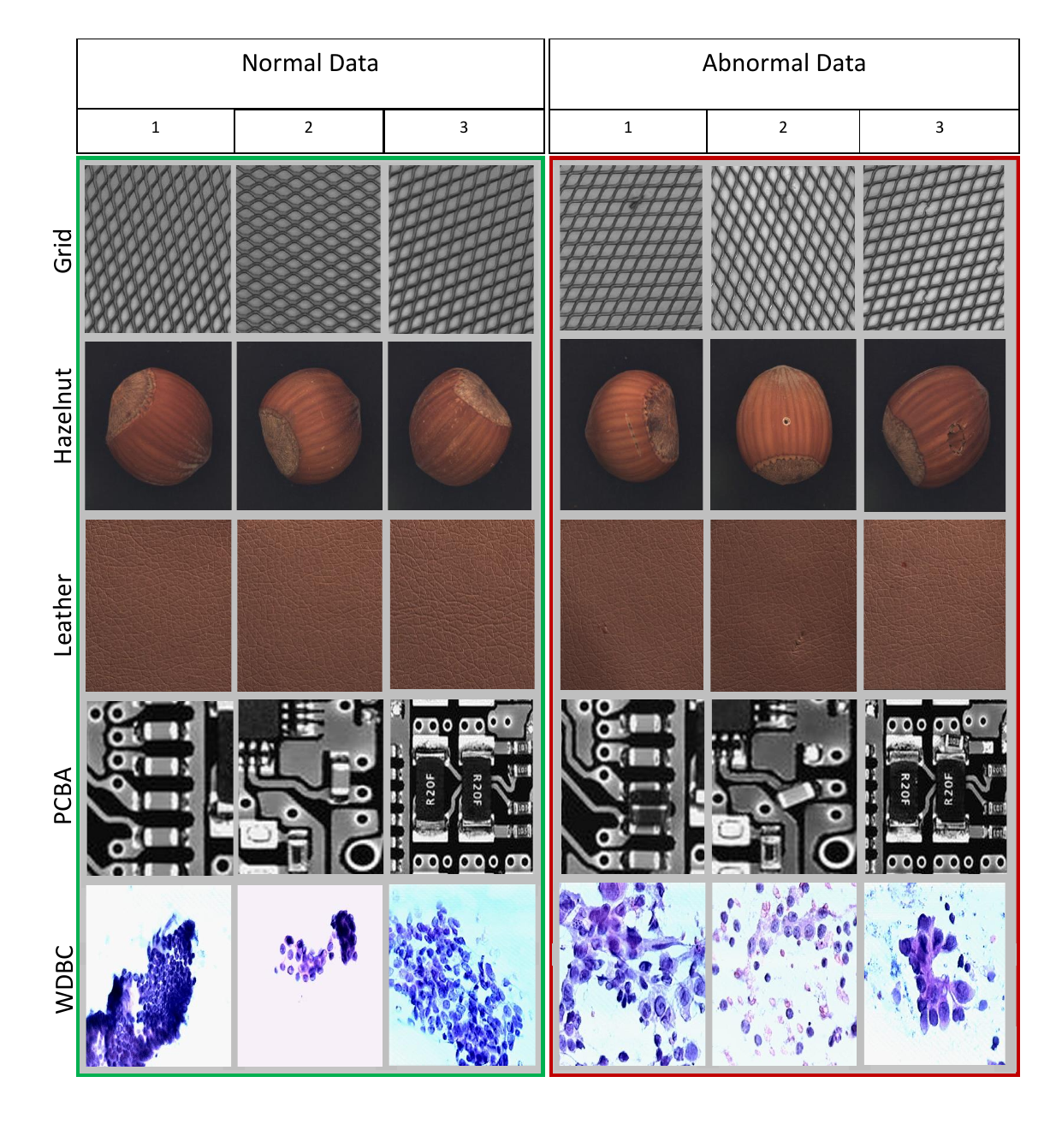}
\caption{Three examples of normal (left green-framed part) and abnormal images (right red-framed part) for the all the datasets coming from image databases. One can observe the very slight difference between the two classes.}\label{Examples}
\end{figure}

Six datasets have been chosen to assess the proposed method. Some of them are linked to a critical application (industrial, medical or cybersecurity), motivating the necessity to avoid false negatives that can yield to dramatic consequences. They are all tabular data, even if, among them, the original format was images. This is the case for the 4 industrial datasets, namely the Leather, Hazelnut and Grid ones coming from the public MVTec-AD image database \cite{bergmann2019mvtec}, and the PCBA image dataset from the private industrial partner. Figure~\ref{Examples} shows some negative (normal) and positive (abnormal) images that display the small variations between both classes.
The images have a resolution of $1024 \times 1024$ and show different nature of complexity and anomalies. The Leather and Grid dataset are texture-like images, whereas the Hazelnut and PCBA are object-like ones. Some images are difficult to classify as normal or abnormal, in particular because the anomalies are so small that they seem to be normal variations. This is for example the case when the task is to detect a small scratch on an Hazelnut image, or a missing component on a PCBA image, as shown in the figure.
Since the customized loss is convex and differentiable, we could have connected a Resnet model (for instance) to process images directly, but we let the image task to a dedicated model. This allows to correctly deal with the resolution needs ($1024 \times 1024$) as well as the subtle normal/abnormal variabilities. Therefore, for the industrial datasets, a first step has been performed thanks to the \textit{VQGanoDIP} method \cite{bougaham2023composite}, namely the "normal version" generation with a VQ-GAN model and the metrics collection from these reconstructed images. All the information coming from this input image reconstruction form  a tabular dataset, with the positive and negative instances in row, and the metrics collected in column (664 features of VQ-GAN component losses, pixel-wise reconstruction quality, patch distances between the input and reconstructed images). This represents a proxy of the differences between the images and the normal version of these ones. When images are normal, these differences are small, which is the not the case when images present an anomaly. All the collected metrics forming the tabular data reflect this statement.

Concerning the two other datasets, the Wisconsin Diagnosis Breast Cancer (WDBC) \cite{breast_cancer_wisconsin_(diagnostic)_17} and the Credit Card Fraud Detection \cite{CCF} (CCF) have been chosen. These tabular datasets are widely used in the literature and are critical applications in the scope of our method. The WDBC dataset reports 30 features extracted from breast images (5th row of the Figure~\ref{Examples}) and describing the cell nucleus (e.g., radius, perimeter, symmetry). The CCF dataset consists of 30 features composed of the time elapsed between two transactions, the transaction amount and 28 principal components obtained with PCA from confidential information. It is populated by 284,807 negative examples, but only 492 ones (same number as positives number) have been randomly selected to reduce memory consumption and training time. Table \ref{datasets_table} shows the number of positive and negative instances for each datasets. One can observe that 4 datasets are imbalanced (Grid, Hazelnut, Leather, WDBC), with a majority of negative data.

\begin{table*}[h]
    \renewcommand{\arraystretch}{1}
    \begin{minipage}{\textwidth} 
    \caption{Number of positive and negative instances for each datasets.}\label{datasets_table}
    {\setlength{\tabcolsep}{0pt}
    \begin{center}
        \begin{tabular*}{\textwidth}{@{\extracolsep{\fill}}lcccc@{\extracolsep{\fill}}}
            \toprule%
            \textit{Dataset} & \textit{\# Negative} & \textit{\# Positive}
            \\\cmidrule{1-1}\cmidrule{2-2}\cmidrule{3-3}%
            CCF & 492 & 492\\
            Grid & 142 & 57\\
            Hazelnut & 215 & 70\\
            Leather & 138 & 92\\
            PCBA & 174 & 174\\
            WDBC & 357 & 212\\
            
            \bottomrule%
        \end{tabular*}
    \end{center}}
    \end{minipage}
\end{table*}

\subsection{Experimental Setup}
\label{Experimental Setup}



The \textit{tapAUC} method implementation comes with some hyperparameter values to chose. A grid search is performed to select the best ones, maximizing the TPR @ FPR $\leq$ 50\%. This 50\% FPR threshold is arbitrary but sufficient to remove all cases where TPR could be 100\% because of a classifier judging all cases as positive, and scoping the false alarms to a reasonable value. The hyperparameters are the following: the total epoch number for training (60, 200, 500), the warmup period (25\%, 50\%, 75\% of the total epoch number), the margin that separates the positive and negative scores $\gamma$ (0.1, 0.3, 0.5, 0.7, 1) and the ratio of negative instances $\alpha$ to select after the warmup period (the single or 5\%, 10\%, 25\%, 50\%  highest scores of the negative instances). The optimizer is Adam with learning rate 0.01. The binary classification model is a fully connected network. It has an input layer with a number of neurons corresponding to the dataset features count, followed by a hidden layer of half the first layer size and an output layer of one neuron terminated by a sigmoïd function. The activation functions are ReLU, batch normalization is used and a dropout of 0.5 is set to avoid overfitting.

To compare the method performance, we ran the same experiments for 4 other methods, namely a traditional binary cross entropy loss (BCE), a squared hinge loss approximating the AUC ROC metric (similar to our method except the partial selection of the negatives), and the one-way KL-DRO estimator (OPAUC) and two-way (TPAUC) partial AUC method proposed in \cite{zhu2022auc}. A grid search has also been used to search the best hyperparameters. For the BCE loss, the total epoch number evaluated is 200 or 500. This is also the case for the squared hinge loss method, as well as the margin $\gamma$ to be 0.1, 0.3, 0.5, 0.7 or 1. For the OPAUC and TPAUC methods, some hyperparameters advised by the authors are kept (total epoch number is 60, $\tau$ is 1), and other ones are grid-searched. They are the margin $\gamma$ (0.5, 0.7 or 1), $\lambda$ (0.1, 1 or 10), and $\gamma$ for OPAUC or $\gamma_0$ and $\gamma_1$ for TPAUC (0.5 or 0.9).

For each dataset, a 5-fold stratified cross validation technique is performed for 5 different repetitions. A post-processing step is executed, in order to drop the constant and correlated features, and to scale their value between 0 and 1. For each fold splits and repetitions, a new seed is given to apply randomization. All experiments have been performed on an Intel i7 CPU and nVidia RTX A5000 GPU, with Python 3.8, Cuda 11.2 and Pytorch 1.10.

\subsection{Results}
\label{Results}

Subsection \ref{Experimental Setup} described the procedure chosen to evaluate the proposed method, and its competitors. The average accuracy, TPR and FPR are reported in Table \ref{metric_table}. This table results from the validation set of the stratified cross validation method, and for the 5 different repetitions. Five different losses are evaluated, being the binary cross entropy, the squared hinge loss, the one-way partial AUC, the two-way partial AUC and our \textit{tapAUC}. Hyperparameters that give the best TPR @ FPR $\leq$ 50\% are chosen, while considering the ZFN threshold.

\begin{table}[h]
\renewcommand{\arraystretch}{1.2}
\centering
\caption{Accuracy, TPR and FPR values by datasets and methods, based on the best TPR @ FPR $\leq$ 50\% from the grid search. Values are expressed in \%, and are averaged by each validation set of the 5-fold cross validation procedure, for 5 different runs. The $\uparrow$ sign means the highest the best, unlike the $\downarrow$ sign means the lowest the best. Bold values are the best ones of each column. Cells with gray background for CCF with BCE and Squared Hinge Loss indicate FPR $>$ 50\%, and are provided for information only. }

\label{metric_table}
\resizebox{\textwidth}{!}{
\begin{tabular}{@{}l*{15}{c}@{}}
\toprule
Method & \multicolumn{3}{c}{BCE} & \multicolumn{3}{c}{Squared Hinge Loss} & \multicolumn{3}{c}{OPAUC} & \multicolumn{3}{c}{TPAUC} & \multicolumn{3}{c}{tapAUC (ours)} \\
\cmidrule{1-1} \cmidrule(lr){2-4} \cmidrule(lr){5-7} \cmidrule(lr){8-10} \cmidrule(lr){11-13} \cmidrule(lr){14-16}
Metric & ACC$\uparrow$ & TPR$\uparrow$ & FPR$\downarrow$ & ACC$\uparrow$ & TPR$\uparrow$ & FPR$\downarrow$ & ACC$\uparrow$ & TPR$\uparrow$ & FPR$\downarrow$ & ACC$\uparrow$ & TPR$\uparrow$ & FPR$\downarrow$ & ACC$\uparrow$ & TPR$\uparrow$ & FPR$\downarrow$ \\
\midrule
CCF & \cellcolor{gray!25}51.89 & \cellcolor{gray!25}99.92 & \cellcolor{gray!25}96.14 & \cellcolor{gray!25}70.21 & \cellcolor{gray!25}98.29 & \cellcolor{gray!25}57.89 & \textbf{86.26} & 96.14 & \textbf{23.62} & 86.2 & 96.06 & 23.66 & 79.69 & \textbf{97.28} & 37.89\\
Grid & 96.27 & 87.36 & \textbf{0.14} & 91.77 & 89.09 & 7.18 & 96.58 & 89.18 & 0.42 & 96.28 & 87.73 & 0.28 & \textbf{96.77} & \textbf{89.42} & 0.28 \\
Hazelnut & \textbf{92.56} & 75.14 & 1.77 & 90.18 & 78.86 & 6.14 & 92.07 & 70.29 & 0.84 & 92.21 & 69.14 & \textbf{0.28} & 85.68 & \textbf{79.14} & 12.19 \\
Leather & 87.48 & 72.62 & \textbf{2.61} & 80.17 & 93.66 & 28.84 & \textbf{89.74} & 81.72 & 4.93 & 89.3 & 79.09 & 3.91 & 85.39 & \textbf{94.39} & 20.58 \\
PCBA & 79.3 & 84.9 & 26.32 & 86.18 & 90.22 & 17.93 & 92.3 & 93.9 & 9.31 & \textbf{93.61} & 92.96 & \textbf{5.75} & 86.95 & \textbf{95.61} & 21.72 \\
WDBC & 81.97 & 99.15 & 28.24 & 81.25 & 98.86 & 29.19 & 92.66 & 97.92 & 10.48 & \textbf{94.62} & 97.83 & \textbf{7.28} & 80.95 & \textbf{99.25} & 29.92 \\

\midrule
MEAN & 81.58 & 86.52 & 25.87 & 83.29 & 91.5 & 24.53 & 91.6 & 88.19 & 8.27 & \textbf{92.04} & 87.14 & \textbf{6.86} & 85.91 & \textbf{92.52} & 20.43 \\
\bottomrule
\end{tabular}
}
\end{table}

\section{Discussion}
\label{Discussion}

The experiments performed in Section \ref{Evaluation} show results that are discussed in this section. 
The first insight of interest in our study is the TPR under a 50\% FPR threshold. This FPR threshold is chosen to bound the false positive alarms at a reasonable value, convenient for a real-word application. For all the datasets, we can see from Table \ref{metric_table} that our \textit{tapAUC} method brings the highest TPR with a mean of 92.52\%, being around 4.3\% and 5.4\% better than OPAUC and TPAUC respectively. This represents an improvement that the business in the field will benefit, by dealing with a smaller ratio of missed detection of anomalies. 
In more detail, the Hazelnut dataset gives the smallest TPR (79.14\%) while WDBC gives the largest one (99.25\%). However, this is not specific to \textit{tapAUC} because all the methods respectively struggle and excel with these two datasets. We can therefore conclude that the Hazelnut dataset is the less expressive in terms of anomaly patterns, whereas the WDBC dataset shows more differences between the benign (negative) and malign (positive) class. Another observation is that results of OPAUC and TPAUC are very close to each other, as well as our \textit{tapAUC} and a simple squared hinge loss. This makes sense because these two pairs of methods are similar to each other and only small changes occur in the subregion targeted. 

For the critical applications our method aims for, the good TPR results at a limited FPR is the priority. Even if it comes with lower accuracy of FPR results, the TPR brings more trustworthiness when the goal is to maximize the detection of the positive instances. In terms of accuracy and FPR, we indeed notice from Table \ref{metric_table} that the methods TPAUC and OPAUC have better results and thus raise much less false alarms. Indeed, our method yields an average of 20.43\% FPR and 85.91\% accuracy whereas, for TPAUC and OPAUC, these values drop to 6.86\% and 8.27\% for the FPR, and jump to 92.04\% and 91.06\% for the accuracy. This is the trade-off to pay for a highly sensitive anomaly detection tool.
In the detail, the methods OPAUC and TPAUC also show similar results of FPR, whereas, this time, it is not the case for the squared hinge loss and our \textit{tapAUC} that have a better FPR. This reflects a better understanding of the negative class for our method, able to limit their misclassification. Indeed, the dynamic focus on the challenging negative instances during training helps to mitigate the false positive, being the cornerstone of our proposal. Finally, one can also notice that for the CCF dataset, the binary cross entropy and the squared hinge losses are not considered as competitors, because they are not able to reach an FPR below 50\%, which is not convenient for real-world business situations.

The observation for OPAUC, TPAUC and \textit{tapAUC} can be explained by the different method strategies. Even if the focus is on the same region of the partial AUC, the way how it is performed is different. Indeed, OPAUC targets the left side of the curve being the lowest FPR possible, and TPAUC does the same but with an additional constraint on the highest TPR possible. However, unlike our method, the training step does not consider any specific threshold where only the smallest optimizable portion is set. For \textit{tapAUC}, this threshold is dynamically calculated so that it targets very precisely the smallest region that prevents the TPR to be 100\% at the lowest FPR possible. While other methods perform this strategy roughly, by associating weights to the bounded region, ours ends up with a dynamic optimization target in the subregion of interest, guiding the classifier to its best FPR at 100\% TPR.

For \textit{tapAUC}, the TPR of Table \ref{metric_table} shows a relatively high value for the different datasets, proving its ability to generalize with unseen data. However, despite these good results, the initial claim was to build a classifier dedicated to reach all the positives well classified. In order to get closer to a more trustworthy tool, an operator solicitation is needed to judge difficult instances that are uncertain, including the few positives misclassified. An uncertainty interval is therefore defined, corresponding to the worst positive score below the threshold as a lower bound, and the threshold itself as an upper bound. All instance scores that fall into this interval is considered as uncertain, and an operator needs to confirm or infirm the classifier decision. Indeed, the closer the anomaly score is to the threshold, the greater the uncertainty. At the opposite, the scores of the negatives far below the threshold are those with the largest confidence, as well as the ones of the positives far above. This human-in-the-loop step helps getting close towards a zero false negative method, but requires a small amount of manual control. We reported in Table \ref{Uncertainty} the lower bound values averaged across all the validation sets of the 5 repetitions, as well as the number of data to manually check it represents and the positives concerned (useful checks), by dataset. 
\begin{table*}[h]
    \renewcommand{\arraystretch}{1}
    \begin{minipage}{\textwidth} 
    \caption{Lower bound, manual and useful checks (in \%) averaged across all the validation sets and repetitions, by dataset. The uncertainty interval is defined as [Threshold-Lower Bound : Threshold]. A mean of 14.71\% of the instances have to be manually checked to guarantee no false negatives, which represents 3.61\% of the positives.}\label{Uncertainty}
    {\setlength{\tabcolsep}{0pt}
    \begin{center}
        \begin{tabular*}{\textwidth}{@{\extracolsep{\fill}}lccccc@{\extracolsep{\fill}}}
            \toprule%
            \textit{Dataset} & \textit{Lower Bound} & \textit{Manual Checks (\%)} & \textit{Useful  Checks (\%)}
            \\\cmidrule{1-1}\cmidrule{2-2}\cmidrule{3-3}\cmidrule{4-4}%
            CCF & 0.1698 & 12.64 & 1.38\\
            Grid & 0.0122 & 13.97 & 4.21\\
            Hazelnut & 0.0173 & 20.84 & 10.86\\
            Leather & 0.0226 & 7.91  & 2.61\\
            PCBA & 0.0078 & 4.77 & 1.95\\
            WDBC & 0.2984 & 28.12 & 0.66\\
            \toprule
            MEAN & 0.088 & 14.71 & 3.61\\            
            \bottomrule%
        \end{tabular*}
    \end{center}}
    \end{minipage}
\end{table*}
This report shows, in average, how the uncertainty interval allows to catch the few positive instances. 14.71\% of the data have to be manually checked, where 3.61\% are useful because positive. This analysis shows that a small amount of manual check brings more confidence, even for the more difficult positive data that the classifier struggles to catch on the validation set.

\section{Conclusion}
\label{Conclusion}

In this work, a trustworthy anomaly detection method called \textit{tapAUC} is proposed, based on an approximated partial AUC (pAUC) loss, and a dedicated algorithm that trains a binary classifier. The main objective is to tackle the trustworthiness constraint required for critical applications such in the industrial, medical or cybersecurity domains. Indeed, these businesses need to detect the positive (abnormal) instances with a high degree of confidence, with a reasonable ratio of false alarms, unlike other non-critical applications that only aim to optimize the accuracy. This requirement is a key characteristic for real-world businesses that want to adopt AI techniques in their daily concerns. It brings a contribution towards the Ethics Guidelines for Trustworthy AI asking to improve explainable AI (XAI) techniques for such organizations. To do so, the model loss is arranged to optimize a pAUC through a squared hinge loss applied to a subset of the positive and the challenging negative instances. This mechanism helps to focus on the specific region where the True Positive Rate (TPR) is just below 100\%. Iteration after iteration, the model is optimized in this region that dynamically changes, and the model gets iteratively optimized by minimizing the False Positive Rate (FPR). At the end, selecting the threshold that generates zero false negatives (ZFN) under limited false positives (LFP) gives a TPR improvement of 4.3\% at a FPR cost of 12.2\% compared to other state-of-the-art methods. This represents a TPR of 92.52\% at a 20.43\% FPR. These results are the validation set metrics average of a stratified 5-fold cross validation technique, performed for 5 repetitions and across 6 datasets. Then, an uncertainty interval is build in order to get closer to a more trustworthy tool. A human-in-the-loop strategy is required for an additional manual check, when the instance prediction score falls close to the threshold. The experiments show that an average of 14.71\% of manual checks are necessary to catch the few misclassified positives.

As future works, these promising results open new research questions. A first one concerns the training strategy. Indeed, the iterative and dynamic focus on a specific region makes the optimization non-smooth and somehow difficult to converge, because the challenging negative instances could change during training. It could be interesting to study how a different approach could smoothen these changes, in order to improve the training step. Another direction concerns the end-to-end learning with images. This study is tailored for tabular data and treat image datasets with the statistics coming from another dedicated method (\textit{VQGanoDIP} \cite{bougaham2023composite}). This is motivated by the $1024 \times 1024$ resolution needed, and the complex discrimination observed at the pixel level. A way to get rid of this limitation is to split the full images into small patches ($224 \times 224$ each for example), and use a Resnet model to extract features in the patches that would replace the tabular statistics data. This would enable a more straightforward learning with eventual better classification performances. Finally, the uncertainty interval strategy still shows possibilities of improvement in terms of building this score-based confidence.

\bibliographystyle{elsarticle-num}
\bibliography{bibliography.bib}






\end{document}